\documentclass[11pt]{article}
\usepackage[margin=1in]{geometry}
\usepackage{times}
\usepackage{natbib}
\setcitestyle{round}

\usepackage{amsmath,amsfonts,bm}

\def\eqref#1{equation~\ref{#1}}

\def\1{\bm{1}}

\DeclareMathAlphabet{\mathsfit}{\encodingdefault}{\sfdefault}{m}{sl}
\SetMathAlphabet{\mathsfit}{bold}{\encodingdefault}{\sfdefault}{bx}{n}

\usepackage{hyperref}
\usepackage{url}
\usepackage{graphicx}
\usepackage{booktabs}
\usepackage{threeparttable}
\usepackage{xspace}
\newcommand{\CLADDER}{CL\textsc{adder}\xspace}

\title{Evaluating Causal Reasoning in LLMs:\\A Case Study in the Challenges\\of Procedural Generation}

\author{%
Justin Clarke \quad Isha Karn \quad Patrick Murray\\[0.4em]
KJ Hill \quad David Jensen\\[0.65em]
\normalsize University of Massachusetts Amherst\\[0.35em]
\small \texttt{jbclarke@umass.edu}, \texttt{ikarn@umass.edu}, \texttt{psmurray@umass.edu}\\
\small \texttt{kjhill424@gmail.com}, \texttt{jensen@umass.edu}
}
\date{}

\begin{document}

\maketitle

\begin{abstract}

    Large language models (LLMs) appear to represent and reason about a large range of phenomena. In particular, recent studies have examined whether LLMs can accurately reason about causation. One of the most commonly used benchmarks for evaluating formal causal reasoning in LLMs is \CLADDER, a framework for procedural generation of benchmark data sets that was released in 2023. Experiments with \CLADDER appear to indicate that LLMs can perform formal causal reasoning in a large percentage of generated examples. In this paper, we examine what can be reliably inferred when LLMs perform well on test sets produced by \CLADDER. We identify several examples of superficial and highly predictive patterns that the \CLADDER generator embeds in data sets it generates. We find that even simple models such as random forests perform well on \CLADDER-generated data sets after training on a sample of generated data. We also find that the performance of many commercial LLMs on \CLADDER test sets is extremely brittle and that removing large sections of the text that are critical for formal causal reasoning has almost no effect on performance. Given these results, the most likely hypothesis for the strong performance of state-of-the-art LLMs on \CLADDER is that they are exploiting superficial patterns learned from exposure to training data from the generator. Any inference that LLMs are exhibiting more complex causal reasoning must rule out this hypothesis.
    
\end{abstract}

\section{Introduction}

One of the most important and challenging questions about large language models (LLMs) is the extent to which these models exhibit complex reasoning. On the one hand, LLMs regularly respond to prompts in ways that appear to require accurate reasoning about logic, mathematics, physics, and causation, as well as knowledge about the natural, physical, and social worlds \citep{hendrycks2021measuring, wei2022chain, chen2021evaluating, kiciman2024causal}. On the other hand, much of this apparent knowledge and reasoning could be recalled verbatim from the massive amounts of data used to train LLMs, as well as some lightweight taxonomic knowledge gleaned from how humans produce natural language \citep{carlini2022quantifying, collins2022structured}. In short, researchers have sought to determine where LLMs sit between the extremes of artificial general intelligence and stochastic parrots \citep{zevcevic2023causal}.

\begin{figure}[!htbp]
    \centering
    \includegraphics[
        width=0.8\linewidth,
        trim={0 0.2cm 0 0.2cm},
        clip
    ]{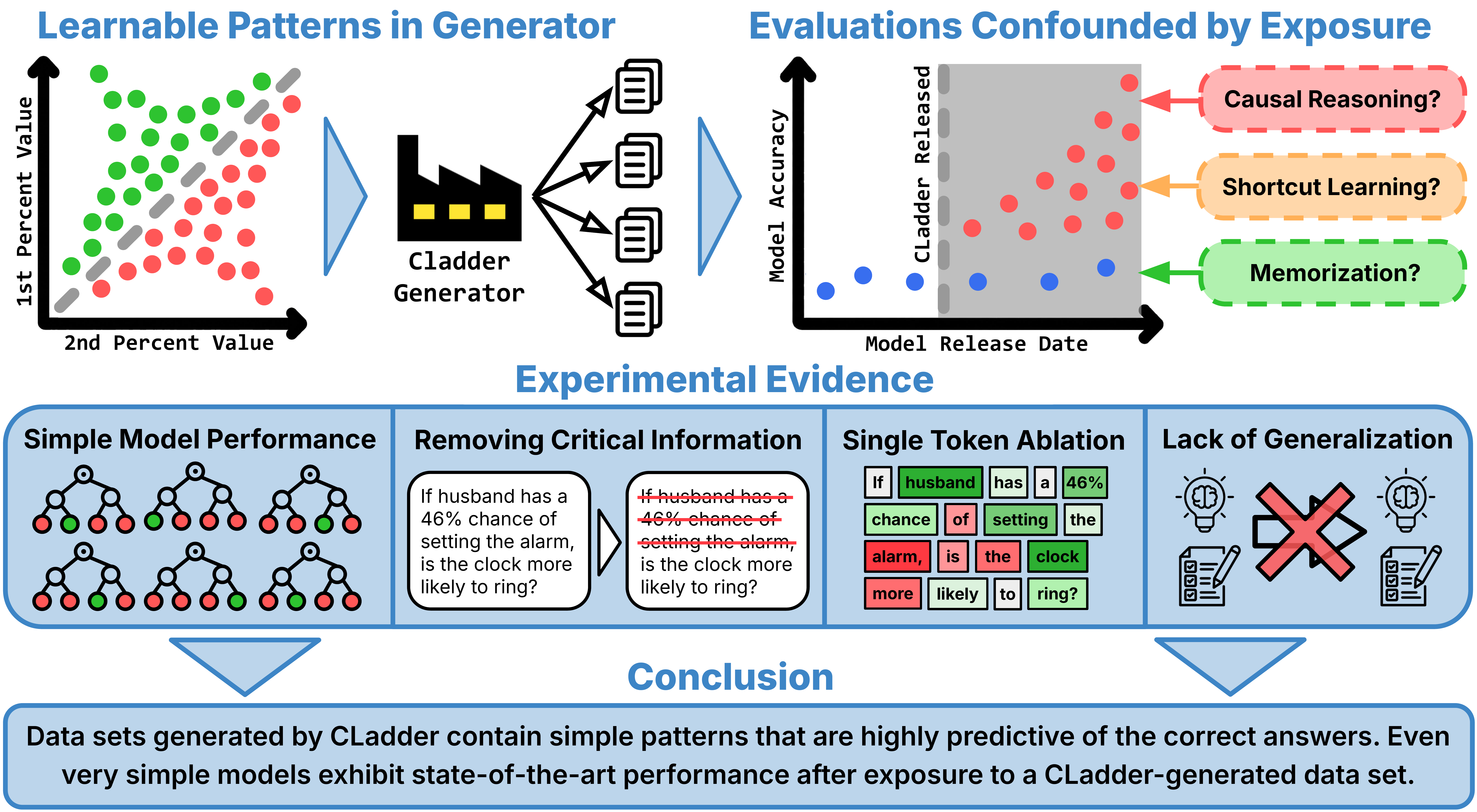}
    \caption{This paper examines how and why LLMs perform well on \CLADDER, a dynamic benchmark for evaluating causal reasoning, and what can be inferred from the results.}
    \label{fig:intro_graphic}
\end{figure}

A major challenge in resolving this question has concerned evaluation — how to create large sets of questions and (correct) answers to evaluate specific types of reasoning. Early evaluations often repurposed existing questions originally designed to assess human knowledge and reasoning, including college entrance examinations, intelligence tests, and textbook exercises \citep[e.g., ][]{achiam2023gpt, hendrycks2021measuring}. However, because many such benchmarks were publicly available before contemporary LLMs were trained, researchers raised concerns that their content may have appeared in model training corpora and been memorized by LLMs \citep{xu2024benchmark, sainz2023nlp}. These concerns are particularly difficult to resolve for closed models whose training data are not publicly documented \citep{deng2024investigating}. Subsequent analyses have found evidence of benchmark-specific overfitting and possible memorization in both open and closed models. \citep[e.g., ][]{zhang2024carefulexaminationlargelanguage}. As one researcher has quipped: “Pretraining on the test set is all you need” \citep{schaeffer2023pretraining}.

Given the problems with static benchmark datasets, an alternative approach to evaluating LLM reasoning has gained prominence: \textit{procedural generation}. With procedural generation, test sets are generated algorithmically rather than being generated manually or gathered from the real world \citep{chen2025benchmarking,risi2020increasing}. In the context of evaluating LLM reasoning, procedural generation allows new test sets to be generated as needed, making it highly unlikely that test set instances could have been memorized \citep{, zhu2024dyval, white2025livebench}. Such “dynamic” test sets appear to overcome some of the most obvious problems associated with earlier “static” test sets. Procedural generation is best suited for reasoning tasks that have been sufficiently formalized such that questions can be randomly sampled from an extremely large (potentially infinite) space and their correct answers can be computed or verified automatically \citep{wei-etal-2025-satbench}. Fortunately, many reasoning tasks have well-established formal theories that allow for the construction of procedural generators. These include mathematics, statistics, logic, and (most importantly for this paper) causal reasoning \citep[e.g.,][]{saxtonanalysing, wei-etal-2025-satbench, jin2023cladder}.

In this work, we examine \CLADDER, a particularly prominent and important procedural generator for evaluating the causal reasoning abilities of LLMs. Specifically, we focus on a single research question: \textbf{\textit{What can be reliably inferred about the reasoning ability of an LLM when it performs well on a \CLADDER-generated test set?}} We focus on this question because a large range of papers have implicitly or explicitly concluded that high performance on \CLADDER-generated test sets implies that a given LLM has learned to perform formal causal inference, and that such reasoning capabilities will generalize to other scenarios. In contrast, our findings strongly indicate that such inferences are unwarranted. A far more likely conclusion is that high-performing models have been exposed to one or more \CLADDER data sets and that the models have learned a set of superficial patterns that allow for high performance. While the ability to learn such patterns is impressive, it is not indicative of formal causal reasoning, and such performance is unlikely to generalize beyond data sets generated by \CLADDER.

As summarized in Figure \ref{fig:intro_graphic}, we show that:

\begin{itemize}

\item \textbf{\CLADDER embeds superficial predictive patterns in the data it produces} --- We show that \CLADDER's data generating process explicitly embeds superficial patterns in every data set that it produces. These patterns are simple and predictive of correct answers.

\item \textbf{These patterns are learnable from exposure to a single \CLADDER data set} --- We show that a variety of simple and complex models can learn these superficial patterns and perform well on subsequent test sets. Simple models such as random forests can learn them when given superficial features derived from the queries. Open-weight LLMs can also learn these predictive patterns from Fine-tuning on \CLADDER-generated queries and answers.

\item \textbf{Exposure to \CLADDER datasets is likely for many modern LLMs} --- \CLADDER-generated data has been popular and has proliferated widely since the paper's initial publication. \CLADDER data is likely to be included in the training corpus of many LLMs. 

\item \textbf{Seemingly essential information is not required for high performance} --- Some information in \CLADDER queries is known to be essential for formal causal reasoning. We show that removing such information does not substantially impair LLM performance.

\item \textbf{Seemingly incidental information is critical to model performance} --- We show that omitting seemingly incidental tokens from specific \CLADDER queries can dramatically affect the answers provided by LLMs.

\item \textbf{High-performance on \CLADDER does not generalize} --- Models that achieve high accuracy on \CLADDER perform no better than random on CounterBench, a recently released benchmark for causal reasoning with a very similar overall design and phrasing of causal queries.

\end{itemize}

\section{\CLADDER: Background, Design, and Prior Results}

Causal reasoning is often identified as one of the most important and complex forms of reasoning that humans do, and thus causal reasoning has become an explicit target of efforts to evaluate the wider reasoning capabilities of LLMs. Several prominent researchers have explicitly identified causal reasoning as a key gap in the reasoning capabilities of LLMs \citep{zhang2023understanding, wu2024causality}. 

Early attempts to empirically evaluate causal reasoning in LLMs concentrated on commonsense causality, assessing whether models correctly assess everyday causal knowledge expressed in natural language \citep{roemmele2011choice}. However, such evaluations conflate causal \textit{knowledge} with causal \textit{reasoning}, since a model could answer correctly by manipulating linguistic patterns rather than reasoning about the abstract causal concepts expressed in that language. Surveys of this work \citep[e.g., ][]{wu2024causality} note that while models can generate causal language that sounds accurate, they fall short of genuine causal reasoning. 

The introduction of the \CLADDER benchmark \citep{jin2023cladder} was a substantial milestone in empirical evaluations of causal reasoning in LLMs. \CLADDER leverages a formal theory of causal reasoning \citep{Pearl2009} to create a procedural generator of question-answer pairs that range across ten distinct types of causal reasoning (average treatment effect, backdoor adjustment, collider bias, correlation, deterministic counterfactual, effect of treatment on the treated, explaining away, marginal effect, natural direct effect, and natural indirect effect) and that span Pearl's three rungs of causation (association, intervention, and counterfactual reasoning). 

\subsection{What is \CLADDER?}
\label{sec:what_is_cladder}

\CLADDER is a procedural generator of question-answer pairs. Questions are generated by randomly sampling from several sets of data structures that completely specify a causal reasoning task (e.g., problem type, graphical model, probability values, etc.) and then converting this information into a query in natural language. Correct answers to the query are also produced automatically by construction, based on the formal data structures. All queries have binary (yes/no) answers. Full details can be found in the original paper \citep{jin2023cladder}.

Such a procedural generation approach would appear to entirely escape the problems of prior ``static'' test sets that could have appeared in the training data of an LLM. Consequently, the results in the original paper (and subsequent papers that have used \CLADDER) have far more credibility than prior work that used static test sets. It is worth noting that, in addition to the \CLADDER generator, \citeauthor{jin2023cladder} released several demonstration test sets that illustrate the ability of the generator to produce data with different properties (e.g., use of nonsense words in place of common nouns and causal dependencies that defied ``commonsense'' expectations). The full demonstration dataset contains 41,748 queries. 

To avoid memorization issues, we generated a new dataset for our experiments with 9866 unique samples that includes all the combinations of options available using the standard implementation of the \CLADDER generator. The dataset contains 1200 queries for each query type, unless there are fewer than 1200 unique possibilities, in which case we include them all. See Appendix \ref{app:generated_dataset} for further details on our generated dataset. 

\subsection{Impact of \CLADDER}
The \CLADDER paper had a strong effect on both the methods and the conclusions of the AI research community. In less than three years since its publication, the paper has been cited over 380 times, and \CLADDER data has been used in more than 20 other empirical evaluations of LLM reasoning \citep[e.g., ][]{yu2025causaleval, he2026uncovering, momente2025triangulating, mao2026recurrent, schrader2024quite}. In some cases, those studies used the \CLADDER generator. In others, however, the original demonstration test sets were used as static benchmarks.

The results in the original paper and subsequent papers that used \CLADDER have been widely cited as evidence that LLMs can perform effective causal reasoning. It is worth noting that \citeauthor{jin2023cladder} were quite cautious in the original paper, saying that ``our work suggests that LLMs may still be far from reasoning reliably about causality.'' That said, they also claim that some LLM prompting methods presented in the paper ``elicit sound causal reasoning in LLMs and help them solve challenging causality questions.'' Subsequent authors have made substantially stronger claims based on results from \CLADDER in the original and subsequent papers \citep[e.g., ][]{gui2025mitigatingspuriouscorrelationsllms, chen2026can}.

Only a few papers have attempted to understand the underlying source of the observed performance in ways that could clarify what can be inferred from high performance. \citeauthor{jin2023cladder} themselves attempted to evaluate the degree to which surface linguistic information in queries affected the answers provided by LLMs. They substituted nonsense words for nouns in causal queries and observed essentially no change in performance. Similarly, \citeauthor{pacchiardi2024leaving} analyzed a variety of LLM reasoning benchmarks for surface linguistic features that predict correct answers, and conclude that \CLADDER test sets are among the least susceptible to learning of generator-specific shortcuts.

\section{Experiments and Key Findings}
In this section, we report the findings from analyses and experiments that inform our view on what can be inferred when a model accurately answers queries in \CLADDER-generated data sets.

\subsection{Highly Predictive Patterns are Directly Embedded by the Generator}

The methods used to generate \CLADDER queries introduce multiple patterns into the data that can be learned and exploited by a model. The data generator produces each query by combining elements from different templates depending on the specified variables and causal graph. This method of generating queries leads to predictable elements that are repeated in any data produced by the generator, such as the structure and specific phrasing of the queries (see Appendix \ref{app:sources-of-determinism} for additional details). In many cases, these patterns provide shortcuts that allow a model to correctly answer the queries without the need for any causal reasoning. 

\begin{figure}[!htbp]
    \centering
    \includegraphics[
        width=1\linewidth,
        trim={0 0.2cm 0 0.2cm},
        clip
    ]{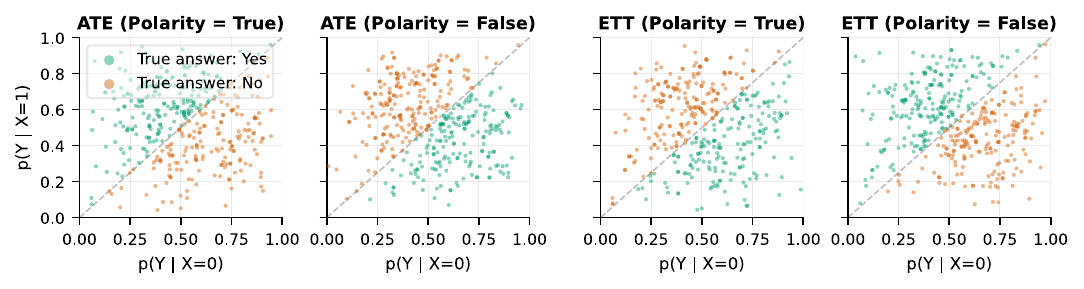}
    \caption{Scatter plots of the two key conditional probabilities provided in average treatment effect (ATE) and effect of treatment on the treated (ETT) queries, colored by correct answer. Within each polarity condition, yes and no answers are nearly linearly separable. A model that learns this simple comparison rule can achieve near-perfect accuracy on these query types without any formal causal reasoning.}
    \label{fig:ATE_first_vs_second_prob}
\end{figure}

For example, after providing the graph structure, many of the average treatment effect queries have the form "$P(X=1|Y=0)=p_1, P(X=1|Y=1)=p_2.$ Will $do(Y=1)$ increase the chance of $X=1$?" This query can be trivially solved by comparing $p_1$ and $p_2$ and checking the polarity of the query (i.e. whether it is phrased as "increase" or "decrease"). Once that pattern is apparent, other elements of the query, including the structure of the graph and names of the variables, are no longer necessary. In Figure \ref{fig:ATE_first_vs_second_prob}, we plot the first probability given in the average treatment effect queries versus the second given probability. We can see that for each polarity the queries can be classified nearly perfectly using only the two probability values. This is not an isolated case: A very similar pattern can be observed in queries that ask about the effect of treatment on the treated. We note that some of the examples shown in Figure \ref{fig:ATE_first_vs_second_prob} have as many as seven probability values provided in the query, even though this highly predictive pattern uses only the first two.

The patterns above are fairly easy to identify, but we have identified other patterns in \CLADDER-generated data that are far more subtle and difficult to detect by inspection alone. Without a way to mitigate this issue, there is a risk that any model exposed to the data during training may learn superficial shortcuts associated with the data generation procedure. This will likely lead to strong performance on the \CLADDER datasets but poor generalization.

\subsection{Simple models can learn the predictive patterns}

We have shown that predictive patterns exist within \CLADDER data, but can these patterns be learned by models and, if so, how well do they work? To answer these questions, we created two sets of superficial features derived directly from the query text: a small set of surface-level features that characterize the query text, as well as a fixed-length BERT embedding  \citep{devlin2019bert} of each query. The BERT embeddings do not represent the probabilities that appear in the query, so they are appended to the embedding (see Appendix \ref{app:random_forest_classifier} for feature descriptions). 

\begin{figure}[!htbp]
    \centering
    \includegraphics[width=1\linewidth,
        trim={0.25cm 0.28cm 0.25cm 0.18cm},
        clip]{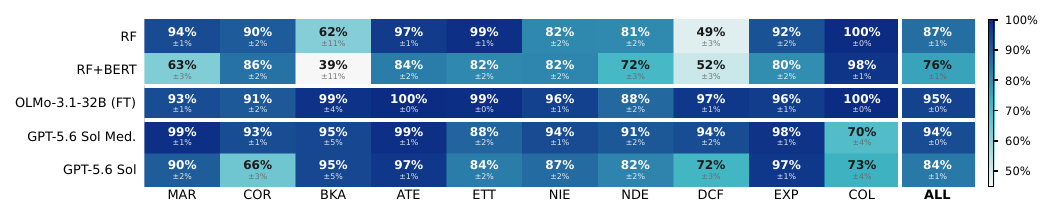}
    \caption{Model accuracies by query type. The first two rows show accuracy by query type on our generated \CLADDER test set for a random forest classifier trained on superficial features derived from the query text or BERT embeddings. We report the mean accuracy over 10-fold cross-validation. The third row shows the performance of OLMo-3.1-32B after fine-tuning on \CLADDER. We also show gpt-5.6-sol with and without reasoning for comparison.}
    \label{fig:random_forest_results}
\end{figure}

We used each set of features to learn a random forest classifier that predicts the binary answer for each query. The cross-validated accuracy estimates for each feature set, disaggregated by query type, are shown in Figure \ref{fig:random_forest_results}. For some query types (e.g., collider bias and marginal), the random forest classifier was able to predict the answer with very high accuracy (100\% and 94\%, respectively). Averaged across all query types, the classifier achieved a cross-validated accuracy of 87\%, roughly comparable to the accuracy achieved by current language models.  

To explore the effect of \CLADDER exposure on LLMs, we also investigate whether supervised fine-tuning on \CLADDER queries produces significant improvements in performance.  We select four instruction-tuned models for which prior exposure to \CLADDER is unlikely based on their release dates or the availability of information about their training data: OLMo-3.1-32B-Instruct, OLMo-3-7B-Instruct, Llama-3.1-8B-Instruct, and Qwen2.5-3B-Instruct. Consistent with limited or no prior exposure, zero-shot performance on the generated queries (n = 9,866) is close to chance for all four models, with accuracies of 63.9 $\pm$ 0.9 \%, 55.3 $\pm$ 1.0, 55.7 $\pm$ 1.0 and 52.1 $\pm$ 1.0, respectively. 

When trained on 2,000 labeled examples, all fine-tuned models exceeded 80\% overall accuracy.  When trained on 16,000 examples,  every model exceeded 94\% overall accuracy. These results, shown in Appendix \ref{app:finetuning}, indicate that all models improve at approximately the same rate, which is not what we would expect from models of such different sizes (Qwen2.5-3B vs OLMo-3.1-32B). The performance of all the models scales with the number of labeled training examples the model has been exposed to, which is consistent with learning superficial features.

Taken together, these results show that combinations of superficial features are sufficient to perform at state-of-the-art levels on \CLADDER test sets. Even limited exposure to \CLADDER data is sufficient for models to learn generator-specific shortcuts that transfer across \CLADDER datasets. At the very least, these results should create a new, minimum baseline for surprising performance of LLM-based models.

\subsection{Models may have been exposed to \CLADDER data}

The ability of relatively simple models to quickly acquire the knowledge necessary for high performance in \CLADDER casts substantial doubt on the conclusion that evaluations using the benchmark indicate any degree of formal causal reasoning. This is particularly true if datasets from \CLADDER have been included in the training data of LLMs. Can we rule out such exposure?

\CLADDER has been publicly available since its original release.  It has been copied, forked, and redistributed across a large and still-growing number of public repositories. We do not claim that any single evaluation, model, or copy of the dataset is the source of this exposure risk, since any one of these many copies is sufficient to contaminate a model's training data regardless of whether it was ever exposed to the canonical release. That alone is enough to complicate the interpretation of any post-release evaluation's results, because high performance cannot cleanly be attributed to improvements in model reasoning abilities when exposure to the data (in some form, through some channel) cannot be ruled out.

\begin{figure}[!htbp]
    \centering
    \includegraphics[
        width=1\linewidth,
        trim={0.1cm 0.28cm 0.15cm 0.18cm},
        clip
    ]{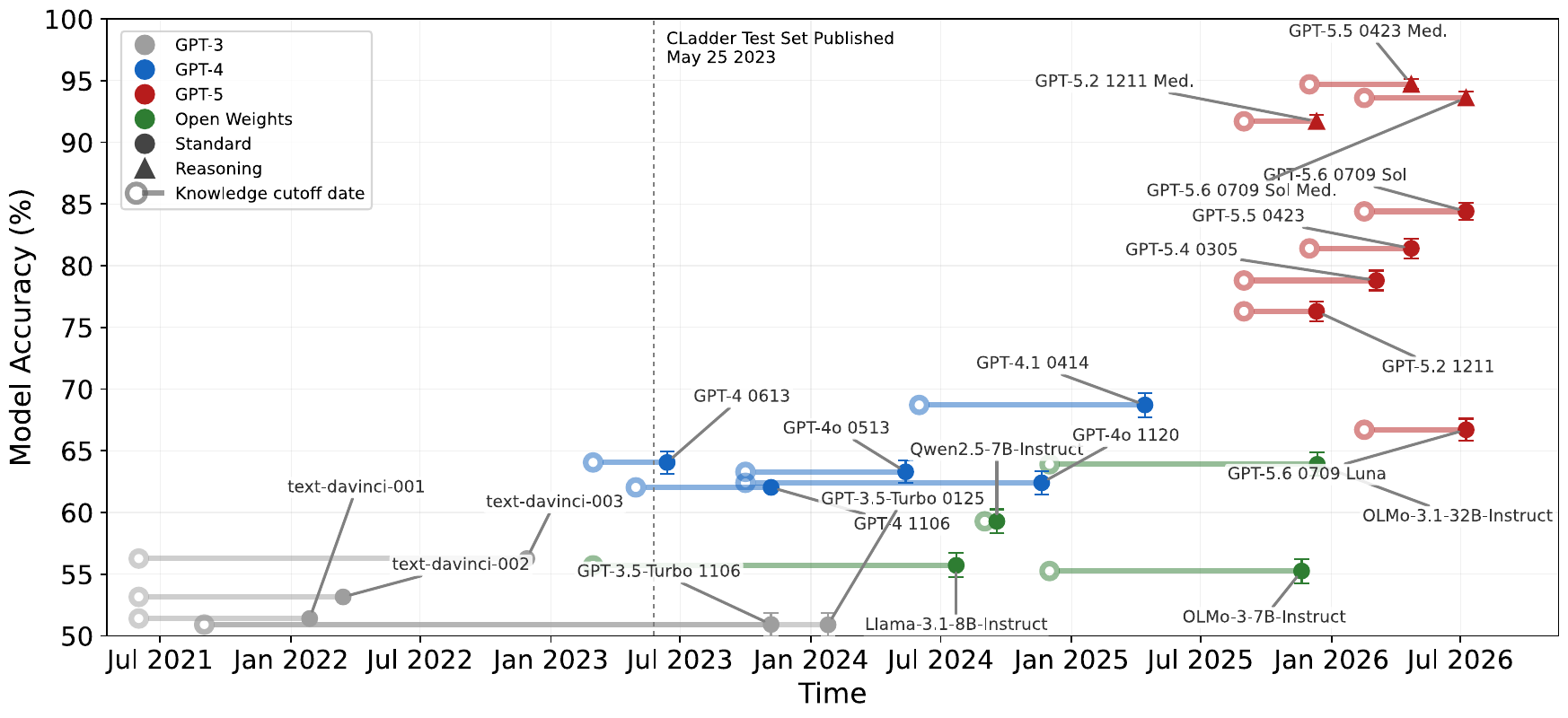}
    \caption{Model accuracy compared to the date of the model's release. Open circles represent the published "knowledge cutoff" dates for each model. The results for the text-davinci models come from the original \CLADDER paper.}  
    \label{fig:accuracy_vs_time}
\end{figure}

We evaluate the performance of several models of varying size and complexity on the dataset we produced using the \CLADDER generator \ref{sec:what_is_cladder}. Figure \ref{fig:accuracy_vs_time} shows the performance of each model versus the date of the model's public release. We observe a clear trend of performance improving over time, with early evaluations finding results slightly above random, and a steady increase in performance leading up to current "reasoning" models, which regularly achieve over 90\% accuracy. But what is the cause of this improvement in performance?

In general, model size and complexity has increased significantly over the time frame shown in Figure \ref{fig:accuracy_vs_time}. However, many of these models were trained after the public release of the \CLADDER generator and example datasets. Without access to the training corpus, there is no way to verify whether the models trained after the dataset's release were exposed during training. In light of our findings above, it appears likely that these models have learned the patterns embedded in the published datasets, confounding any evaluation of their causal reasoning abilities using \CLADDER as a benchmark.

\subsection{Removing essential information has little effect on performance}
\label{sec:model-high-perf-essential}

Above we show that the use of simple features is sufficient to explain high performance on \CLADDER. Next, we take a different approach: We examine whether LLMs still perform well on \CLADDER queries when they lack the information that is necessary for formal causal reasoning. Specifically, we remove key components of the queries that leave them unsolvable by formal causal reasoning, and we find that such ablations produce no substantial drop in LLM performance.

\begin{figure}[!htbp]
    \centering
    \includegraphics[width=1\linewidth,
        trim={0 0.4cm 0 0.18cm},
        clip]
    {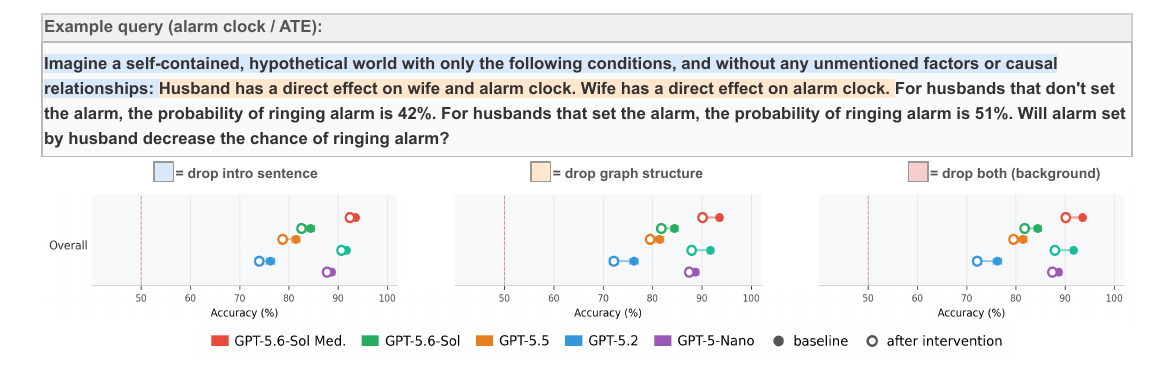}
    \caption{Model accuracy under interventions that remove elements of the queries that are critical for formal causal reasoning. A full query-wise breakdown in Appendix \ref{app:query-wise-acc}. The 95\% Wilson confidence intervals are not visible in this plot due to large sample size.}
    \label{fig:intervention_1B}
\end{figure}

Each query begins with a statement establishing a set of background conditions and specifying the structure of the causal graph used to generate the query (see Figure \ref{fig:intervention_1B} for an example). Together, these elements are necessary to make the causal query ``identifiable'' (answerable with a unique quantitative solution).  Our experiments remove each of these elements individually, and then together. In all cases, we observe minimal reductions in accuracy. Across five evaluated frontier models, these interventions resulted in reductions in overall accuracy of only 0.6 to 4.2 percentage points relative to the baseline performance on the unaltered queries.

Next, we perform a set of even more aggressive interventions, resulting in queries where it is impossible to calculate the answer with the available information. We begin by removing only the first marginal or conditional probability value provided in the query, followed by removing only the second value provided. Finally, we remove all the marginal or conditional probability values that are provided with the query (identified as the "given\_info" in the \CLADDER queries). The result in all cases are queries with insufficient information to calculate the correct answer.  

\begin{figure}[!htbp]
    \centering
    \includegraphics[width=1\linewidth,
        trim={0 0.4cm 0 0.18cm},
        clip]{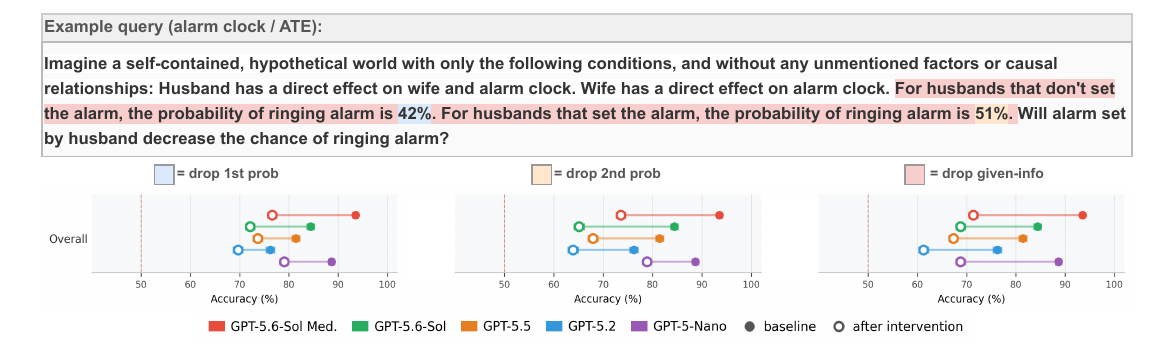}
    \caption{Model accuracy under interventions that remove elements of the queries that make it impossible to calculate the answer. Full query-wise breakdown in Appendix \ref{app:query-wise-acc}. The 95\% Wilson confidence intervals are not visible in this plot due to large sample size.}
    \label{fig:intervention_2B}
\end{figure}

These interventions, shown in Figure \ref{fig:intervention_2B}, result in a larger reduction in accuracy than those in Figure \ref{fig:intervention_1B}, which makes sense due to the critical nature of the information being removed. However, we would expect performance to collapse to 50\% in all three cases, as models should not be able to perform better than random chance. Instead, we see the accuracy reduced, but still remaining well above 50\%. Across all three versions of the intervention, the models perform  11 to 21 percentage points above random. 

Taken together, these results indicate that the models are unlikely to be performing formal causal reasoning.  A model that was doing so should not perform any better than random after such critical information is removed. In fact, these results show that large sections of the text that are critical for formal causal reasoning actually have a negligible effect on model performance. Instead, the models appear to be exploiting other shortcuts to correctly answer the queries, including ones that do not rely on elements of the query that formal causal reasoning methods would require. 

\subsection{Seemingly incidental information greatly affects performance}
\label{sec:heatmap_experiments}

The results above indicate that models largely ignore information that is essential for formal causal reasoning. That raises a natural question: What information \textit{do} models use?  In the next set of experiments, we aimed to localize the elements of the query that have the largest impact on model performance. We establish the model's baseline accuracy for each query, and then we re-process the query with a single token removed and observe any change in performance. The models we tested proved to be very brittle under these types of interventions, and highly sensitive to the presence or absence of specific tokens. We often see a large change in performance when tokens are removed that are incidental to a human's ability to solve the problem, and we find extreme variations in both baseline performance and sensitivity to individual tokens for nearly identical queries.

\begin{figure}[!htbp]
    \centering
    \includegraphics[width=1\linewidth,
        trim={0 0.25cm 0 0.3cm},
        clip]{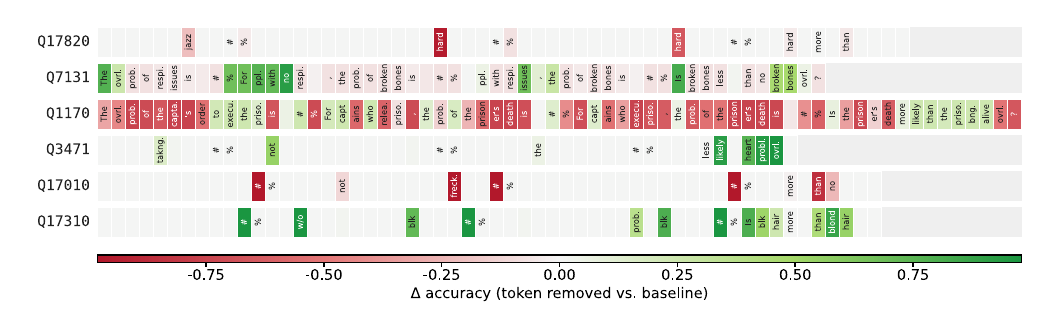}
    \caption{Token ablation results for a set of marginal queries on gpt-5.2. Each row corresponds to a single prompt, and each cell represents one token. Red cells denote tokens whose removal \textit{reduces} model performance, while green cells denote tokens whose removal \textit{increases} performance. Numerical tokens (\#), percentage tokens (\%), and the polarity indicator are always labeled. Otherwise, labels are shown only for tokens exceeding a salience threshold ($|\Delta| \geq 0.04$).} 
    \label{fig:top-12-ablation}
\end{figure}

We find a surprising amount of variation in which tokens affect model performance, even between queries of the same type. Figure \ref{fig:top-12-ablation} shows the model's sensitivity to the removal of the token in each position for a set of queries drawn from the `` marginal'' queries. We run each query 100 times to establish a baseline accuracy, then 100 times for each token that is removed. We can see that despite being nearly identical in difficulty and appearance from a human's perspective, there is a significant variation in the response to the removal of each token. There are several tokens whose removal produces a complete collapse in accuracy ($\Delta = -1.0$), suggesting the model relies heavily on specific tokens. See Appendix \ref{app:token_ablation} for additional examples.

Looking closely at Figure \ref{fig:top-12-ablation} we can see the surprising variation we find in the model's response to nearly identical queries. For the first query, labeled $Q17820$, where the baseline accuracy is 100\%, removing the single token ``hard'' in the compound word ``hard-working'' can reliably flip the model's answer such that its accuracy on the query is approximately 0\%. The second query in the figure, labeled $Q7131$, is nearly identical aside from the variable names. Except in this case we observe a baseline accuracy of only 8.0\%, even though these queries would be of equal difficulty for a human. Additionally, for the query with low baseline performance we find there are many tokens whose removal results in a significant increase in performance to nearly 100\% accuracy. And again, we see that removing tokens that would appear inconsequential to a human seems to have a large impact on the model's performance. 

Variation in the model's response to similar queries, as well as the surprising impact of individual tokens on model performance, are both strong indicators that the models are exploiting superficial patterns to solve the queries, and that disrupting these patterns can cause dramatic shifts in performance.

\subsection{Model Performance on \CLADDER Does Not Generalize}

The experiments described above establish that high performance on \CLADDER datasets is achievable through mechanisms that do not require formal causal reasoning. If models were performing formal causal reasoning, or even exploiting broadly generalizable patterns, we would expect these abilities to generalize to structurally similar causal reasoning tasks. If instead their performance is dependent on patterns that are specific to \CLADDER's data generation process, we would not expect those patterns to generalize to an independent benchmark.

We evaluate several frontier models on CounterBench \citep{chen2026counterbench}, a recently released benchmark for counterfactual causal reasoning that uses abstract variable names and was independently produced. Table \ref{tab:counterbench} in Appendix \ref{app:counterbench} shows the accuracy for each model on CounterBench. All frontier models score near chance on CounterBench (48.6\% – 52.2\%). The gap between \CLADDER and CounterBench performance for different models ranges from 29 to 44 percentage points.

Surprisingly, the random forest classifier trained exclusively on superficial features derived from query text transfers substantially better to CounterBench, achieving 79.1\%, as shown in Table \ref{tab:counterbench}. It performed substantially above every frontier LLM we tested, suggesting that the frontier LLMs are not relying on the same transferable combination of features captured by the random forests or the fine-tuned models.

\section{Discussion and Conclusion}

The converging lines of evidence in Section 3 strongly suggest that high performance on \CLADDER does not reliably imply that LLMs are performing some type of formal causal reasoning. Instead, exposure to \CLADDER data and superficial pattern learning seems much more likely. Clearly, this has direct implications for what we should conclude from the results of prior evaluations that employed \CLADDER data sets and how the generator should be used in future studies.

More broadly, our results illustrate a limitation of procedural generation as a strategy for evaluating LLM reasoning. Many of the patterns in \CLADDER are not easily detected by inspection, and they appear to be an artifact of the generation process itself rather than resulting from design flaws that could be easily fixed. The combination of procedural generation and public examples of generated data may be incompatible with evaluating powerful machine learning models. While such an approach largely escapes the problem of a model memorizing training data that will also appear as test data, it does not escape the problem of a model learning a \textit{generator’s} intrinsic patterns.

This suggests at least three possible ways forward. First, in theory, a procedural generator could be constructed that is so faithful to the intended domain that its only embedded patterns are those that will generalize to real world systems. This seems unlikely, unless real-world systems are directly used to generate data. Second, a less faithful procedural generator could be constructed and then withheld from public view. Without public access to either the generator itself or data from the generator, its intrinsic patterns could not easily be learned. This strategy has already been taken in other evaluations of causal inference methods that have been somewhat successful (e.g., \cite{dorie2019automated}). Third, if neither of these approaches is practical, then LLMs should always be compared to a baseline of a simpler model directly trained on the relevant public data set. This will, at least, prevent some types of unwarranted conclusions about the complex reasoning capabilities of LLMs.

\subsection*{AI use statement}

We used LLMs during the preparation of this paper for code assistance. All experimental design, analysis, and interpretation of the results were done by the authors. No text in the paper was generated by AI.

\subsection*{Ethics statement}

This work uses two existing public benchmark datasets (\CLADDER and CounterBench) and a newly generated dataset produced with \CLADDER's own procedural generator. No data involves human subjects, personally identifiable information, or sensitive content.

\subsection*{Reproducibility statement}

The codebase for this study includes intervention scripts, simple classifier code, and fine-tuning configurations. Generated parameters and fine-tuning hyperparameters are documented in the appendix.

\bibliographystyle{plainnat} 
\bibliography{references}

@article{jin2023cladder,
  title={Cladder: Assessing causal reasoning in language models},
  author={Jin, Zhijing and Chen, Yuen and Leeb, Felix and Gresele, Luigi and Kamal, Ojasv and Lyu, Zhiheng and Blin, Kevin and Gonzalez Adauto, Fernando and Kleiman-Weiner, Max and Sachan, Mrinmaya and others},
  journal={Advances in Neural Information Processing Systems},
  volume={36},
  pages={31038--31065},
  year={2023}
}

@article{dorie2019automated,
  title={Automated versus do-it-yourself methods for causal inference: Lessons learned from a data analysis competition},
  author={Dorie, Vincent and Hill, Jennifer and Shalit, Uri and Scott, Marc and Cervone, Dan},
  journal={Statistical Science},
  volume={34},
  number={1},
  pages={43–68},
  year={2019}
}

@inproceedings{devlin2019bert,
  title={Bert: Pre-training of deep bidirectional transformers for language understanding},
  author={Devlin, Jacob and Chang, Ming-Wei and Lee, Kenton and Toutanova, Kristina},
  booktitle={Proceedings of the 2019 conference of the North American chapter of the association for computational linguistics: human language technologies, volume 1 (long and short papers)},
  pages={4171--4186},
  year={2019}
}

@article{pacchiardi2024leaving,
  title={Leaving the barn door open for Clever Hans: Simple features predict LLM benchmark answers},
  author={Pacchiardi, Lorenzo and Tesic, Marko and Cheke, Lucy G and Hern{\'a}ndez-Orallo, Jos{\'e}},
  journal={arXiv preprint arXiv:2410.11672},
  year={2024}
}

@inproceedings{chen2025benchmarking,
  title={Benchmarking large language models under data contamination: A survey from static to dynamic evaluation},
  author={Chen, Simin and Chen, Yiming and Li, Zexin and Jiang, Yifan and Wan, Zhongwei and He, Yixin and Ran, Dezhi and Gu, Tianle and Li, Haizhou and Xie, Tao and others},
  booktitle={Proceedings of the 2025 Conference on Empirical Methods in Natural Language Processing},
  pages={10091--10109},
  year={2025}
}

@article{risi2020increasing,
  title={Increasing generality in machine learning through procedural content generation},
  author={Risi, Sebastian and Togelius, Julian},
  journal={Nature Machine Intelligence},
  volume={2},
  number={8},
  pages={428--436},
  year={2020},
  publisher={Nature Publishing Group UK London}
}

@article{achiam2023gpt,
  title={Gpt-4 technical report},
  author={Achiam, Josh and Adler, Steven and Agarwal, Sandhini and Ahmad, Lama and Akkaya, Ilge and Aleman, Florencia Leoni and Almeida, Diogo and Altenschmidt, Janko and Altman, Sam and Anadkat, Shyamal and others},
  journal={arXiv preprint arXiv:2303.08774},
  year={2023}
}

@article{xu2024benchmark,
  title={Benchmark data contamination of large language models: A survey},
  author={Xu, Cheng and Guan, Shuhao and Greene, Derek and Kechadi, M and others},
  journal={arXiv preprint arXiv:2406.04244},
  year={2024}
}

@BOOK{Pearl2009,
  title     = "Causality",
  author    = "Pearl, Judea",
  publisher = "Cambridge University Press",
  month     =  sep,
  year      =  2009,
  address   = "Cambridge"
}

@misc{lora,
      title={LoRA: Low-Rank Adaptation of Large Language Models}, 
      author={Edward J. Hu and Yelong Shen and Phillip Wallis and Zeyuan Allen-Zhu and Yuanzhi Li and Shean Wang and Lu Wang and Weizhu Chen},
      year={2021},
      eprint={2106.09685},
      archivePrefix={arXiv},
      primaryClass={cs.CL},
      url={https://arxiv.org/abs/2106.09685}, 
}

@misc{qlora,
      title={QLoRA: Efficient Finetuning of Quantized LLMs}, 
      author={Tim Dettmers and Artidoro Pagnoni and Ari Holtzman and Luke Zettlemoyer},
      year={2023},
      eprint={2305.14314},
      archivePrefix={arXiv},
      primaryClass={cs.LG},
      url={https://arxiv.org/abs/2305.14314}, 
}

@article{Geirhos_2020,
   title={Shortcut learning in deep neural networks},
   volume={2},
   ISSN={2522-5839},
   url={http://dx.doi.org/10.1038/s42256-020-00257-z},
   DOI={10.1038/s42256-020-00257-z},
   number={11},
   journal={Nature Machine Intelligence},
   publisher={Springer Science and Business Media LLC},
   author={Geirhos, Robert and Jacobsen, Jörn-Henrik and Michaelis, Claudio and Zemel, Richard and Brendel, Wieland and Bethge, Matthias and Wichmann, Felix A.},
   year={2020},
   month=Nov, pages={665–673} 
}

@misc{zhang2024carefulexaminationlargelanguage,
      title={A Careful Examination of Large Language Model Performance on Grade School Arithmetic}, 
      author={Hugh Zhang and Jeff Da and Dean Lee and Vaughn Robinson and Catherine Wu and Will Song and Tiffany Zhao and Pranav Raja and Charlotte Zhuang and Dylan Slack and Qin Lyu and Sean Hendryx and Russell Kaplan and Michele Lunati and Summer Yue},
      year={2024},
      eprint={2405.00332},
      archivePrefix={arXiv},
      primaryClass={cs.CL},
      url={https://arxiv.org/abs/2405.00332}, 
}

@inproceedings{mccoy2019right,
  title={Right for the wrong reasons: Diagnosing syntactic heuristics in natural language inference},
  author={McCoy, R Thomas and Pavlick, Ellie and Linzen, Tal},
  booktitle={Proceedings of the 57th annual meeting of the association for computational linguistics},
  pages={3428--3448},
  year={2019}
}

@article{de2025semantic,
  title={Semantic Deception: When Reasoning Models Can't Compute an Addition},
  author={de Leeuw, Nathani{\"e}l and Nahon, Marceau and Reymond, Mathis and Chatila, Raja and Khamassi, Mehdi},
  journal={arXiv preprint arXiv:2512.20812},
  year={2025}
}

@inproceedings{sainz2023nlp,
  title={NLP evaluation in trouble: On the need to measure LLM data contamination for each benchmark},
  author={Sainz, Oscar and Campos, Jon and Garc{\'\i}a-Ferrero, Iker and Etxaniz, Julen and de Lacalle, Oier Lopez and Agirre, Eneko},
  booktitle={Findings of the Association for Computational Linguistics: EMNLP 2023},
  pages={10776--10787},
  year={2023}
}

@article{chang2026right,
  title={Right for the wrong reasons: Epistemic regret minimization for causal rung collapse in llms},
  author={Chang, Edward Y},
  journal={arXiv preprint arXiv:2602.11675},
  year={2026}
}

@article{zevcevic2023causal,
  title={Causal parrots: Large language models may talk causality but are not causal},
  author={Ze{\v{c}}evi{\'c}, Matej and Willig, Moritz and Dhami, Devendra Singh and Kersting, Kristian},
  journal={arXiv preprint arXiv:2308.13067},
  year={2023}
}

@article{chen2026can,
  title={Can Post-Training Transform LLMs into Causal Reasoners?},
  author={Chen, Junqi and Chen, Sirui and Lu, Chaochao},
  journal={arXiv preprint arXiv:2602.06337},
  year={2026}
}

@article{wu2024causality,
  title={Causality for large language models},
  author={Wu, Anpeng and Kuang, Kun and Zhu, Minqin and Wang, Yingrong and Zheng, Yujia and Han, Kairong and Li, Baohong and Chen, Guangyi and Wu, Fei and Zhang, Kun},
  journal={arXiv preprint arXiv:2410.15319},
  year={2024}
}

@article{li2026leveraging,
  title={Leveraging Large Language Models for Causal Discovery: a Constraint-based, Argumentation-driven Approach},
  author={Li, Zihao and Russo, Fabrizio},
  journal={arXiv preprint arXiv:2602.16481},
  year={2026}
}

@article{yamaoka2026linear,
  title={Linear-LLM-SCM: Benchmarking LLMs for Coefficient Elicitation in Linear-Gaussian Causal Models},
  author={Yamaoka, Kanta and Mukherjee, Sumantrak and G{\"a}rtner, Thomas and Selby, David Antony and Konigorski, Stefan and H{\"u}llermeier, Eyke and Bengs, Viktor and Vollmer, Sebastian Josef},
  journal={arXiv preprint arXiv:2602.10282},
  year={2026}
}

@article{geng2026causalt5k,
  title={Causalt5k: Diagnosing and informing refusal for trustworthy causal reasoning of skepticism, sycophancy, detection-correction, and rung collapse},
  author={Geng, Longling and Ouyang, Andy and Wu, Theodore and Barretto, Daphne and Hayes, Matthew John and Cooper, Rachael and Zeng, Yuqiao and Vijay, Sameer and Ancone, Gia and Rai, Ankit and others},
  journal={arXiv preprint arXiv:2602.08939},
  year={2026}
}

@article{maasch2025causalarc,
  title={CausalARC: Abstract Reasoning with Causal World Models},
  author={Maasch, Jacqueline and Kalantari, John and Khezeli, Kia},
  journal={arXiv preprint arXiv:2509.03636},
  year={2025}
}

@article{maasch2025compositional,
  title={Compositional causal reasoning evaluation in language models},
  author={Maasch, Jacqueline RMA and H{\"u}y{\"u}k, Alihan and Xu, Xinnuo and Nori, Aditya V and Gonzalez, Javier},
  journal={arXiv preprint arXiv:2503.04556},
  year={2025}
}

@article{du2025ice,
  title={Ice Cream Doesn't Cause Drowning: Benchmarking LLMs Against Statistical Pitfalls in Causal Inference},
  author={Du, Jin and Chen, Li and Xian, Xun and Luo, An and Tian, Fangqiao and Wang, Ganghua and Doss, Charles and Shen, Xiaotong and Ding, Jie},
  journal={arXiv preprint arXiv:2505.13770},
  year={2025}
}

@article{schaeffer2023pretraining,
  title={Pretraining on the test set is all you need},
  author={Schaeffer, Rylan},
  journal={arXiv preprint arXiv:2309.08632},
  year={2023}
}

@article{zhang2023understanding,
  title={Understanding causality with large language models: Feasibility and opportunities},
  author={Zhang, Cheng and Bauer, Stefan and Bennett, Paul and Gao, Jiangfeng and Gong, Wenbo and Hilmkil, Agrin and Jennings, Joel and Ma, Chao and Minka, Tom and Pawlowski, Nick and others},
  journal={arXiv preprint arXiv:2304.05524},
  year={2023}
}

@article{kiciman2024causal,
  title={Causal reasoning and large language models: Opening a new frontier for causality},
  author={Kiciman, Emre and Ness, Robert and Sharma, Amit and Tan, Chenhao},
  journal={Transactions on Machine Learning Research},
  year={2024}
}

@inproceedings{roemmele2011choice,
  title={Choice of Plausible Alternatives: An Evaluation of Commonsense Causal Reasoning.},
  author={Roemmele, Melissa and Bejan, Cosmin Adrian and Gordon, Andrew S},
  booktitle={AAAI spring symposium: logical formalizations of commonsense reasoning},
  pages={90--95},
  year={2011}
}

@misc{gui2025mitigatingspuriouscorrelationsllms,
      title={Mitigating Spurious Correlations in LLMs via Causality-Aware Post-Training}, 
      author={Shurui Gui and Shuiwang Ji},
      year={2025},
      eprint={2506.09433},
      archivePrefix={arXiv},
      primaryClass={cs.LG},
      url={https://arxiv.org/abs/2506.09433}, 
}

@inproceedings{hendrycks2021measuring,
  title     = {Measuring Massive Multitask Language Understanding},
  author    = {Hendrycks, Dan and Burns, Collin and Basart, Steven and
               Zou, Andy and Mazeika, Mantas and Song, Dawn and
               Steinhardt, Jacob},
  booktitle = {International Conference on Learning Representations},
  year      = {2021}
}

@article{wei2022chain,
  title={Chain-of-thought prompting elicits reasoning in large language models},
  author={Wei, Jason and Wang, Xuezhi and Schuurmans, Dale and Bosma, Maarten and Xia, Fei and Chi, Ed and Le, Quoc V and Zhou, Denny and others},
  journal={Advances in neural information processing systems},
  volume={35},
  pages={24824--24837},
  year={2022}
}

@article{chen2021evaluating,
  title={Evaluating large language models trained on code},
  author={Chen, Mark and Tworek, Jerry and Jun, Heewoo and Yuan, Qiming and Pinto, Henrique Ponde De Oliveira and Kaplan, Jared and Edwards, Harri and Burda, Yuri and Joseph, Nicholas and Brockman, Greg and others},
  journal={arXiv preprint arXiv:2107.03374},
  year={2021}
}

@inproceedings{carlini2022quantifying,
  title={Quantifying memorization across neural language models},
  author={Carlini, Nicholas and Ippolito, Daphne and Jagielski, Matthew and Lee, Katherine and Tramer, Florian and Zhang, Chiyuan},
  booktitle={The Eleventh International Conference on Learning Representations},
  year={2022}
}

@inproceedings{collins2022structured,
  title={Structured, flexible, and robust: benchmarking and improving large language models towards more human-like behavior in out-of-distribution reasoning tasks},
  author={Collins, Katherine M and Wong, Catherine and Feng, Jiahai and Wei, Megan and Tenenbaum, Josh},
  booktitle={Proceedings of the Annual Meeting of the Cognitive Science Society},
  volume={44},
  number={44},
  year={2022}
}

@inproceedings{deng2024investigating,
  title={Investigating data contamination in modern benchmarks for large language models},
  author={Deng, Chunyuan and Zhao, Yilun and Tang, Xiangru and Gerstein, Mark and Cohan, Arman},
  booktitle={Proceedings of the 2024 Conference of the North American Chapter of the Association for Computational Linguistics: Human Language Technologies (Volume 1: Long Papers)},
  pages={8706--8719},
  year={2024}
}

@inproceedings{zhu2024dyval,
  title={Dyval: Dynamic evaluation of large language models for reasoning tasks},
  author={Zhu, Kaijie and Chen, Jiaao and Wang, Jindong and Gong, Neil and Yang, Diyi and Xie, Xing},
  booktitle={International Conference on Learning Representations},
  volume={2024},
  pages={18091--18128},
  year={2024}
}

@inproceedings{white2025livebench,
  title={Livebench: A challenging, contamination-limited llm benchmark},
  author={White, Colin and Dooley, Samuel and Roberts, Manley and Pal, Arka and Feuer, Benjamin and Jain, Siddhartha and Shwartz-Ziv, Ravid and Jain, Neel and Saifullah, Khalid and Dey, Sreemanti and others},
  booktitle={International Conference on Learning Representations},
  volume={2025},
  pages={91595--91631},
  year={2025}
}

@inproceedings{wei-etal-2025-satbench,
    title = "{SATB}ench: Benchmarking {LLM}s' Logical Reasoning via Automated Puzzle Generation from {SAT} Formulas",
    author = "Wei, Anjiang  and
      Wu, Yuheng  and
      Wan, Yingjia  and
      Suresh, Tarun  and
      Tan, Huanmi  and
      Zhou, Zhanke  and
      Koyejo, Sanmi  and
      Wang, Ke  and
      Aiken, Alex",
    editor = "Christodoulopoulos, Christos  and
      Chakraborty, Tanmoy  and
      Rose, Carolyn  and
      Peng, Violet",
    booktitle = "Proceedings of the 2025 Conference on Empirical Methods in Natural Language Processing",
    month = nov,
    year = "2025",
    address = "Suzhou, China",
    publisher = "Association for Computational Linguistics",
    url = "https://aclanthology.org/2025.emnlp-main.1716/",
    doi = "10.18653/v1/2025.emnlp-main.1716",
    pages = "33832--33849",
    ISBN = "979-8-89176-332-6"
}

@inproceedings{saxtonanalysing,
  title={Analysing Mathematical Reasoning Abilities of Neural Models},
  author={Saxton, David and Grefenstette, Edward and Hill, Felix and Kohli, Pushmeet},
  booktitle={International Conference on Learning Representations},
  year=2019
}

@inproceedings{yu2025causaleval,
  title={Causaleval: Towards better causal reasoning in language models},
  author={Yu, Longxuan and Chen, Delin and Xiong, Siheng and Wu, Qingyang and Li, Dawei and Chen, Zhikai and Liu, Xiaoze and Pan, Liangming},
  booktitle={Proceedings of the 2025 Conference of the Nations of the Americas Chapter of the Association for Computational Linguistics: Human Language Technologies (Volume 1: Long Papers)},
  pages={12512--12540},
  year={2025}
}

@inproceedings{he2026uncovering,
    title = "Uncovering Hidden Correctness in {LLM} Causal Reasoning via Symbolic Verification",
    author = "He, Paul  and
      Huang, Yinya  and
      Sachan, Mrinmaya  and
      Jin, Zhijing",
    editor = "Demberg, Vera  and
      Inui, Kentaro  and
      Marquez, Llu{\'i}s",
    booktitle = "Proceedings of the 19th Conference of the {E}uropean Chapter of the {A}ssociation for {C}omputational {L}inguistics (Volume 1: Long Papers)",
    month = mar,
    year = "2026",
    address = "Rabat, Morocco",
    publisher = "Association for Computational Linguistics",
    url = "https://aclanthology.org/2026.eacl-long.56/",
    doi = "10.18653/v1/2026.eacl-long.56",
    pages = "1231--1250",
    ISBN = "979-8-89176-380-7"
}

@inproceedings{momente2025triangulating,
    title = "Triangulating {LLM} Progress through Benchmarks, Games, and Cognitive Tests",
    author = "Moment{\`e}, Filippo  and
      Suglia, Alessandro  and
      Giulianelli, Mario  and
      Ferrari, Ambra  and
      Koller, Alexander  and
      Lemon, Oliver  and
      Schlangen, David  and
      Fern{\'a}ndez, Raquel  and
      Bernardi, Raffaella",
    editor = "Christodoulopoulos, Christos  and
      Chakraborty, Tanmoy  and
      Rose, Carolyn  and
      Peng, Violet",
    booktitle = "Findings of the Association for Computational Linguistics: EMNLP 2025",
    month = nov,
    year = "2025",
    address = "Suzhou, China",
    publisher = "Association for Computational Linguistics",
    url = "https://aclanthology.org/2025.findings-emnlp.1092/",
    doi = "10.18653/v1/2025.findings-emnlp.1092",
    pages = "20051--20072",
    ISBN = "979-8-89176-335-7"
}

@INPROCEEDINGS{mao2026recurrent,
  author={Mao, Zhenjiang and Venkat, Anirudhh},
  booktitle={ICASSP 2026 - 2026 IEEE International Conference on Acoustics, Speech and Signal Processing (ICASSP)}, 
  title={Recurrent Confidence Chain: Temporal-Aware Uncertainty Quantification in Large Language Models}, 
  year={2026},
  volume={},
  number={},
  pages={18562-18566},
  doi={10.1109/ICASSP55912.2026.11461479}}

@inproceedings{schrader2024quite,
  title={QUITE: quantifying uncertainty in natural language text in Bayesian reasoning scenarios},
  author={Schrader, Timo Pierre and Lange, Lukas and Razniewski, Simon and Friedrich, Annemarie},
  booktitle={Proceedings of the 2024 Conference on Empirical Methods in Natural Language Processing},
  pages={2634--2652},
  year={2024}
}

@inproceedings{chen2026counterbench,
  title={CounterBench: Evaluating and improving counterfactual reasoning in large language models},
  author={Chen, Yuefei and Singh, Vivek K and Ma, Jing and Tang, Ruixiang},
  booktitle={Proceedings of the AAAI Conference on Artificial Intelligence},
  volume={40},
  number={36},
  pages={30350--30358},
  year={2026}
}

\newpage
\appendix
\section{Appendix}
\label{app:related-work}

\subsection{Critiques of \CLADDER and Synthetic Benchmarks}

Several recent papers have identified limitations of \CLADDER from directions complementary to ours. \citet{geng2026causalt5k} and \citet{chang2026right} argue that \CLADDER's template-heavy queries have low linguistic diversity and fail to capture failure modes that appear in naturalistic causal judgments, positioning their own benchmarks as testing more realistic reasoning scenarios. \citet{maasch2025causalarc} explicitly cite benchmark leakage as a motivation for designing CausalARC around randomly sampled tasks that cannot be memorized, noting that o4-mini offered unprompted \CLADDER-style outputs suggestive of training exposure. \citet{maasch2025compositional} extend this concern with a compositional framework, finding that only o1 does not degrade with reasoning complexity, which is consistent with our findings that apparent causal competence collapses under targeted perturbations. \citet{du2025ice} show that \CLADDER-style benchmarks do not test statistical pitfalls such as survivorship bias and collider conditioning in realistic scenarios, and find that model conclusions shift based on semantic label choice. 

More recent work has examined specific subproblems: \citet{li2026leveraging} use LLMs for causal discovery via graph structure identification, while \citet{yamaoka2026linear} benchmark coefficient elicitation in linear Gaussian Structural Causal Models. These complementary directions highlight that causal competence is multidimensional and that strong performance on one subproblem does not imply general causal reasoning ability.

\subsection{Shortcut Learning and Benchmark Contamination}
The phenomenon of models achieving high benchmark accuracy through spurious correlations rather than the intended reasoning process is well documented \citep{Geirhos_2020, mccoy2019right}. \citet{chang2026right} formalize this as Rung Collapse --- the systematic substitution of associational reasoning for interventional queries. \citet{de2025semantic} demonstrate analogous surface exploitation in arithmetic, showing that LLMs default to learned semantic associations over abstract reasoning when tested on novel symbol systems. On the contamination side, \citet{zhang2024carefulexaminationlargelanguage} show that models exhibit inflated performance on publicly available math benchmarks relative to held-out equivalents, with accuracy gaps correlating with the probability of generating training examples. \citet{chen2026can} show that fine-tuning on synthetic SCM data produces dramatic accuracy gains similar to our LoRA experiments, but do not investigate whether those gains reflect genuine reasoning or shortcut learning.

\subsection{Generated dataset}
\label{app:generated_dataset}

We generated a \CLADDER dataset using the original codebase of \citet{jin2023cladder}, running \texttt{fig generate} across the configurations (\texttt{final/common} and \texttt{final/det-common}).

This produced 9,866 unique (question, answer) pairs, which we used for all of our experiments. We use this generated split rather than the original v1 release for our evaluation as our test set for the fine-tuning experiment.

\begin{table}[h]
\centering
\small
\begin{tabular}{lrc}
\toprule
\textbf{Query type} & \textbf{Count} & \textbf{Rung} \\
\midrule
Marginal             & 1200 & 1 \\
Correlation          & 1200 & 1 \\
Collider bias        &  432 & 1 \\
Backdoor adjustment  &   74 & 2 \\
ATE                  & 1200 & 2 \\
ETT                  & 1200 & 2 \\
Exp-away             & 1200 & 2 \\
NIE                  & 1200 & 3 \\
NDE                  & 1200 & 3 \\
Det-counterfactual   &  960 & 3 \\
\midrule
\textbf{Total}       & \textbf{9{,}866} & \\
\bottomrule
\end{tabular}
\caption{Query-type distribution in the generated dataset.}
\label{tab:generated_qt}
\end{table}

Query-types vary in count because the underlying causal graph templates vary in richness and would generate duplicates. The original \CLADDER releases included surface-level interventions as a robustness test. We did not apply these during our dataset generation, as we found no significant difference in model performance between these dataset split. 

To confirm our generated data matches the difficulty of the original \CLADDER v1 release, we evaluate gpt-5.5 (medium reasoning) zero-shot on our generated set and each of the four v1 splits. Table \ref{tab:split_comparison} shows nearly identical performance on all splits (all $p > 0.05$, none significant after Bonferroni correction).

\begin{table}[h]
\centering
\caption{Zero-shot GPT-5.5 (medium reasoning) accuracy on our generated dataset (\textsc{Generated}) and each \textsc{CLadder} v1 split individually. No split is significantly harder than the generated dataset ($p$-values from two-proportion $z$-tests, Bonferroni-corrected threshold $p < 0.0125$, none significant).}
\label{tab:split_comparison}
\begin{tabular}{lrrcc}
\toprule
Split & $N$ & Acc.\,(\%) & 95\% CI & $p$-value \\
\midrule
\textit{Generated} & 9,866 & 94.73 & $\pm$0.4 & — \\
Easy & 10,559 & 94.16 & $\pm$0.4 & 0.074 \\
Hard & 10,556 & 94.49 & $\pm$0.4 & 0.443 \\
Non-commonsense & 10,240 & 94.11 & $\pm$0.5 & 0.056 \\
Anti-commonsense & 10,392 & 95.20 & $\pm$0.4 & 0.127 \\
\bottomrule
\end{tabular}
\end{table}

\newpage
\subsection{Model Performance on Generated \CLADDER Data}

\begin{table*}[!htbp]
\centering
\setlength{\tabcolsep}{5pt}
\renewcommand{\arraystretch}{1.1}
\caption{Model accuracy on generated \textsc{Cladder} queries ($\%$, $\pm$ Wilson 95\,\% CI half-width). Bold = best per column. \dag~Back-Adj has $n{=}74$; all other types $n{\geq}432$.}
\label{tab:generated-baselines}
\resizebox{\textwidth}{!}{%
\begin{tabular}{lrrrrrrrrrrr}
\toprule
\textbf{Model} & \textbf{Overall} & \textbf{ATE} & \textbf{MAR} & \textbf{COR} & \textbf{EXP} & \textbf{NDE} & \textbf{DCF} & \textbf{NIE} & \textbf{ETT} & \textbf{COL} & \textbf{BKA} \\
\midrule
\multicolumn{12}{l}{\textit{Non-LLM baselines}} \\[-2pt]
\midrule
RF        & 86.8{\tiny$\pm$0.7} & 96.8{\tiny$\pm$1.0} & 93.8{\tiny$\pm$1.4} & 90.4{\tiny$\pm$1.7} & 92.2{\tiny$\pm$1.5} & 81.2{\tiny$\pm$2.2} & 48.6{\tiny$\pm$3.2} & 81.5{\tiny$\pm$2.2} & 99.0{\tiny$\pm$0.6} & 100.0{\tiny$\pm$0.4} & 62.2{\tiny$\pm$10.8} \\
RF+BERT   & 76.2{\tiny$\pm$0.8} & 83.8{\tiny$\pm$2.1} & 62.7{\tiny$\pm$2.7} & 85.8{\tiny$\pm$2.0} & 80.2{\tiny$\pm$2.3} & 71.5{\tiny$\pm$2.6} & 51.7{\tiny$\pm$3.2} & 81.8{\tiny$\pm$2.2} & 81.6{\tiny$\pm$2.2} & 97.9{\tiny$\pm$1.4} & 39.2{\tiny$\pm$10.9} \\
\midrule
\multicolumn{12}{l}{\textit{Open-weight models}} \\[-2pt]
\midrule
Llama-3.1-8B-Instruct & 55.7{\tiny$\pm$1.0} & 79.5{\tiny$\pm$2.3} & 52.2{\tiny$\pm$2.8} & 46.3{\tiny$\pm$2.8} & 52.8{\tiny$\pm$2.8} & 51.2{\tiny$\pm$2.8} & 61.0{\tiny$\pm$3.1} & 49.0{\tiny$\pm$2.8} & 60.8{\tiny$\pm$2.8} & 39.8{\tiny$\pm$4.6} & 50.0{\tiny$\pm$11.1} \\
Qwen2.5-3B-Instruct & 52.1{\tiny$\pm$1.0} & 87.6{\tiny$\pm$1.9} & 47.8{\tiny$\pm$2.8} & 54.0{\tiny$\pm$2.8} & 15.4{\tiny$\pm$2.0} & 55.4{\tiny$\pm$2.8} & 54.5{\tiny$\pm$3.1} & 54.0{\tiny$\pm$2.8} & 49.8{\tiny$\pm$2.8} & 50.0{\tiny$\pm$4.7} & 50.0{\tiny$\pm$11.1} \\
OLMo-3-7B-Instruct & 55.3{\tiny$\pm$1.0} & 88.3{\tiny$\pm$1.8} & 50.1{\tiny$\pm$2.8} & 61.2{\tiny$\pm$2.8} & 24.0{\tiny$\pm$2.4} & 54.0{\tiny$\pm$2.8} & 54.6{\tiny$\pm$3.1} & 54.0{\tiny$\pm$2.8} & 51.7{\tiny$\pm$2.8} & 66.0{\tiny$\pm$4.5} & 56.8{\tiny$\pm$11.0} \\
OLMo-3.1-32B-Instruct & 63.9{\tiny$\pm$0.9} & 97.3{\tiny$\pm$0.9} & 54.5{\tiny$\pm$2.8} & 53.2{\tiny$\pm$2.8} & 70.1{\tiny$\pm$2.6} & 70.7{\tiny$\pm$2.6} & 63.2{\tiny$\pm$3.0} & 60.2{\tiny$\pm$2.8} & 47.5{\tiny$\pm$2.8} & 50.0{\tiny$\pm$4.7} & 51.4{\tiny$\pm$11.1} \\
\midrule
\multicolumn{12}{l}{\textit{GPT models}} \\[-2pt]
\midrule
\midrule
GPT-3.5-turbo & 55.5{\tiny$\pm$1.0} & 74.9{\tiny$\pm$2.5} & 49.4{\tiny$\pm$2.8} & 53.0{\tiny$\pm$2.8} & 40.4{\tiny$\pm$2.8} & 55.3{\tiny$\pm$2.8} & 52.1{\tiny$\pm$3.2} & 55.4{\tiny$\pm$2.8} & 63.9{\tiny$\pm$2.7} & 55.6{\tiny$\pm$4.7} & 40.5{\tiny$\pm$10.9} \\
GPT-4-0613 & 71.0{\tiny$\pm$0.9} & 98.5{\tiny$\pm$0.7} & 62.9{\tiny$\pm$2.7} & 63.2{\tiny$\pm$2.7} & 72.8{\tiny$\pm$2.5} & 69.5{\tiny$\pm$2.6} & 67.1{\tiny$\pm$3.0} & 63.9{\tiny$\pm$2.7} & 74.3{\tiny$\pm$2.5} & 62.5{\tiny$\pm$4.5} & 37.8{\tiny$\pm$10.8} \\
GPT-4o-2024-11-20 & 70.6{\tiny$\pm$0.9} & 95.6{\tiny$\pm$1.2} & 69.5{\tiny$\pm$2.6} & 57.2{\tiny$\pm$2.8} & 76.4{\tiny$\pm$2.4} & 70.2{\tiny$\pm$2.6} & 61.7{\tiny$\pm$3.1} & 70.8{\tiny$\pm$2.6} & 71.1{\tiny$\pm$2.6} & 51.6{\tiny$\pm$4.7} & 35.1{\tiny$\pm$10.6} \\
GPT-4.1-2025-04-14 & 75.6{\tiny$\pm$0.8} & 94.4{\tiny$\pm$1.3} & 74.2{\tiny$\pm$2.5} & 73.0{\tiny$\pm$2.5} & 87.4{\tiny$\pm$1.9} & 72.0{\tiny$\pm$2.5} & 69.9{\tiny$\pm$2.9} & 71.3{\tiny$\pm$2.6} & 69.3{\tiny$\pm$2.6} & 56.9{\tiny$\pm$4.6} & 62.2{\tiny$\pm$10.8} \\
GPT-5.2 & 76.3{\tiny$\pm$0.8} & 94.2{\tiny$\pm$1.3} & 74.9{\tiny$\pm$2.5} & 66.6{\tiny$\pm$2.7} & 92.3{\tiny$\pm$1.5} & 70.5{\tiny$\pm$2.6} & 62.2{\tiny$\pm$3.1} & 72.9{\tiny$\pm$2.5} & 82.4{\tiny$\pm$2.2} & 50.7{\tiny$\pm$4.7} & 81.1{\tiny$\pm$8.8} \\
GPT-5.2 (med.\ reasoning) & 91.7{\tiny$\pm$0.5} & \textbf{99.6{\tiny$\pm$0.4}} & \textbf{99.1{\tiny$\pm$0.6}} & 94.1{\tiny$\pm$1.3} & \textbf{98.4{\tiny$\pm$0.7}} & 86.8{\tiny$\pm$1.9} & 88.9{\tiny$\pm$2.0} & 89.6{\tiny$\pm$1.7} & \textbf{88.6{\tiny$\pm$1.8}} & 58.6{\tiny$\pm$4.6} & 93.2{\tiny$\pm$6.0} \\
GPT-5.4-2026-03-05 & 78.8{\tiny$\pm$0.8} & 92.8{\tiny$\pm$1.5} & 83.3{\tiny$\pm$2.1} & 77.5{\tiny$\pm$2.4} & 93.5{\tiny$\pm$1.4} & 72.7{\tiny$\pm$2.5} & 64.2{\tiny$\pm$3.0} & 68.8{\tiny$\pm$2.6} & 82.5{\tiny$\pm$2.1} & 56.0{\tiny$\pm$4.7} & 82.4{\tiny$\pm$8.6} \\
GPT-5.5-2026-04-23 & 81.4{\tiny$\pm$0.8} & 96.2{\tiny$\pm$1.1} & 82.8{\tiny$\pm$2.1} & 70.5{\tiny$\pm$2.6} & 97.4{\tiny$\pm$0.9} & 80.2{\tiny$\pm$2.3} & 70.1{\tiny$\pm$2.9} & 79.0{\tiny$\pm$2.3} & 81.6{\tiny$\pm$2.2} & 56.9{\tiny$\pm$4.6} & 82.4{\tiny$\pm$8.6} \\
GPT-5.5 (med.\ reasoning) & \textbf{94.7{\tiny$\pm$0.4}} & \textbf{99.6{\tiny$\pm$0.3}} & 98.9{\tiny$\pm$0.6} & \textbf{94.2{\tiny$\pm$1.3}} & 98.3{\tiny$\pm$0.7} & 89.6{\tiny$\pm$1.7} & \textbf{94.0{\tiny$\pm$1.5}} & 93.3{\tiny$\pm$1.4} & \textbf{88.6{\tiny$\pm$1.8}} & \textbf{94.5{\tiny$\pm$2.1}} & 92.4{\tiny$\pm$5.5} \\
GPT-5.6-luna & 66.7{\tiny$\pm$0.9} & 93.2{\tiny$\pm$1.4} & 66.2{\tiny$\pm$2.7} & 63.7{\tiny$\pm$2.7} & 49.3{\tiny$\pm$2.8} & 65.4{\tiny$\pm$2.7} & 61.3{\tiny$\pm$3.1} & 67.5{\tiny$\pm$2.6} & 73.1{\tiny$\pm$2.5} & 50.0{\tiny$\pm$4.7} & 43.2{\tiny$\pm$11.0} \\
GPT-5.6-sol & 84.4{\tiny$\pm$0.7} & 97.4{\tiny$\pm$0.9} & 90.2{\tiny$\pm$1.7} & 65.6{\tiny$\pm$2.7} & 97.3{\tiny$\pm$0.9} & 82.2{\tiny$\pm$2.2} & 72.2{\tiny$\pm$2.8} & 87.2{\tiny$\pm$1.9} & 84.1{\tiny$\pm$2.1} & 73.1{\tiny$\pm$4.2} & \textbf{94.6{\tiny$\pm$5.5}} \\
GPT-5.6-sol (med.\ reasoning) & 93.5{\tiny$\pm$0.5} & 99.1{\tiny$\pm$0.5} & 98.8{\tiny$\pm$0.6} & 93.2{\tiny$\pm$1.4} & 98.3{\tiny$\pm$0.7} & \textbf{91.1{\tiny$\pm$1.6}} & 93.4{\tiny$\pm$1.6} & \textbf{93.5{\tiny$\pm$1.4}} & 88.3{\tiny$\pm$1.8} & 69.7{\tiny$\pm$4.3} & 92.4{\tiny$\pm$5.5} \\
\bottomrule
\end{tabular}%
}
\end{table*}

\subsection{Fine-tuning Details}
\label{app:finetuning}

We fine-tune four open-weight models ( OLMo-3-7B-Instruct, OLMo-3.1-32B-Instruct, Llama-3.1-8B-Instruct, and Qwen2.5-3B) on the public \CLADDER dataset, stratified by query type at 80/20 split for train and val. We use the generated \CLADDER dataset as our test set. We perform parameter-efficient fine-tuning via LoRA \citep{lora}. To study the effects of training data size, we vary the number of training examples $N \in {50, 200, 500, 1\text{k}, 2\text{k}, 4\text{k}, 8\text{k}, 16\text{k}}$ and evaluate on the held-out test set. We use rank $r = 16$, scaling factor $\alpha=32$, dropout $0.05$, targeting both attention projection layers (\texttt{q\_proj}, \texttt{k\_proj}, \texttt{v\_proj}, \texttt{o\_proj}) and feed-forward layers (\texttt{gate\_proj}, \texttt{up\_proj}, \texttt{down\_proj}), following \citet{qlora}. Models are trained for 5 epochs with AdamW, cosine learning rate $2 \times 10^{-4}$, effective batch size $16$, bf16 precision with gradient checkpointing, and prompt tokens masked to $-100$ so only the answer token contributes to training loss.

\begin{figure}[!htbp]
    \centering
    \includegraphics[
        width=1\linewidth,
        trim={0 0.2cm 0 0.2cm},
        clip
    ]{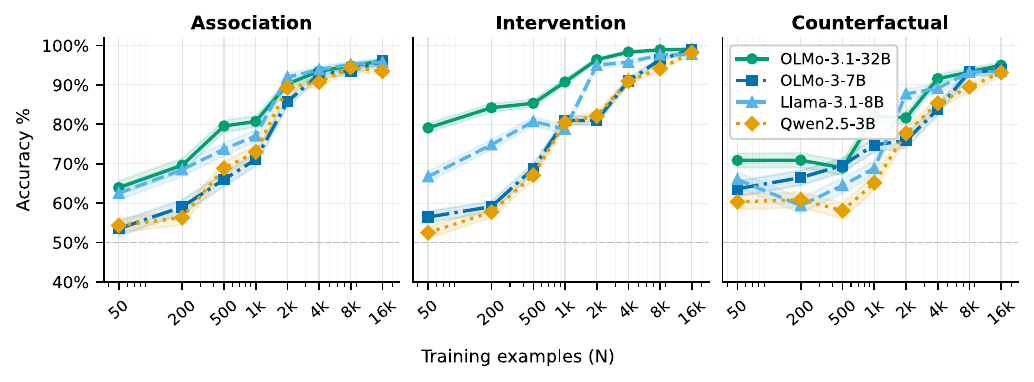}
    \caption{Model performance as a function of the number of publicly-available \CLADDER samples the model is exposed to during fine-tuning.}
    \label{fig:graded-exposure}
\end{figure}

\begin{figure}[!htbp]
    \centering
    \includegraphics[
        width=1\linewidth,
        trim={0 0.2cm 0 0.2cm},
        clip
    ]{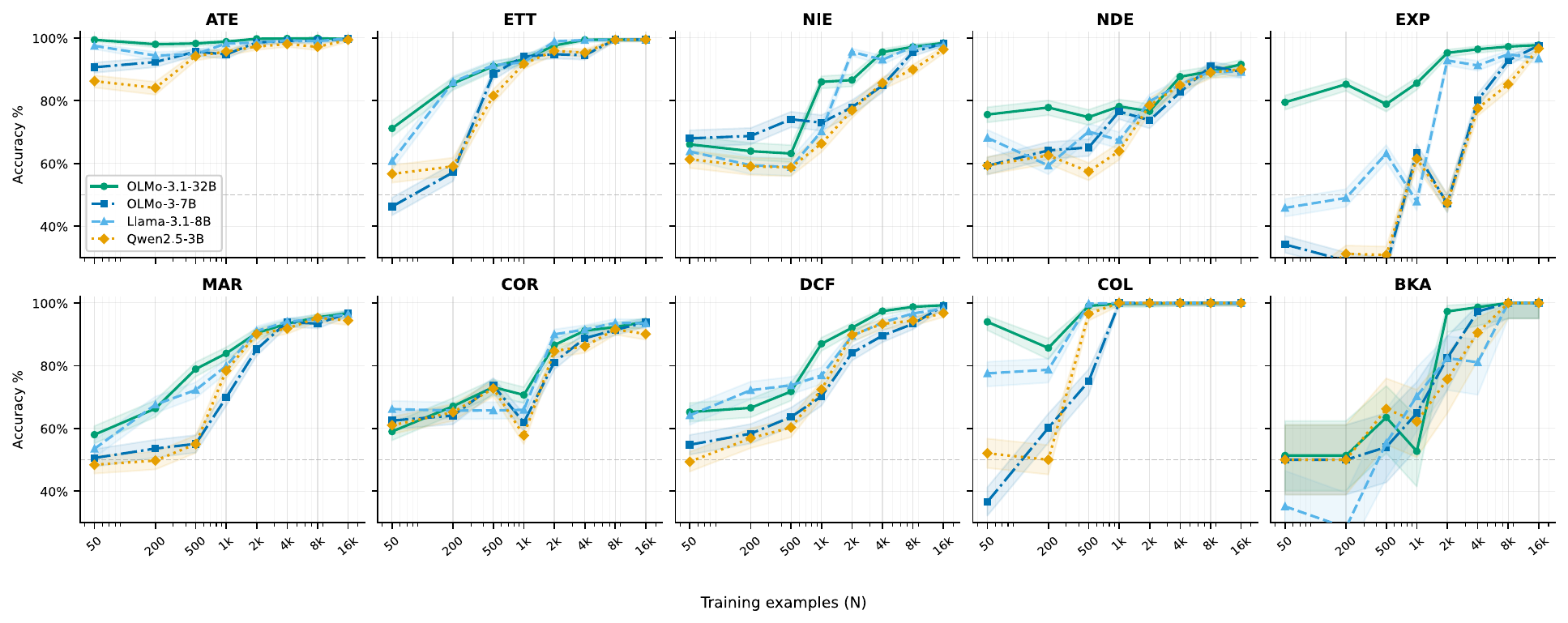}
    \caption{Model performance (per query-type) as a function of the number of publicly-available \CLADDER samples the model is exposed to during fine-tuning.}
    \label{fig:per-querytype-graded-exposure}
\end{figure}

\subsection{Query-wise Accuracy After Performing Interventions}
\label{app:query-wise-acc}

\begin{figure}[!htbp]
    \centering
    \includegraphics[width=1\linewidth]{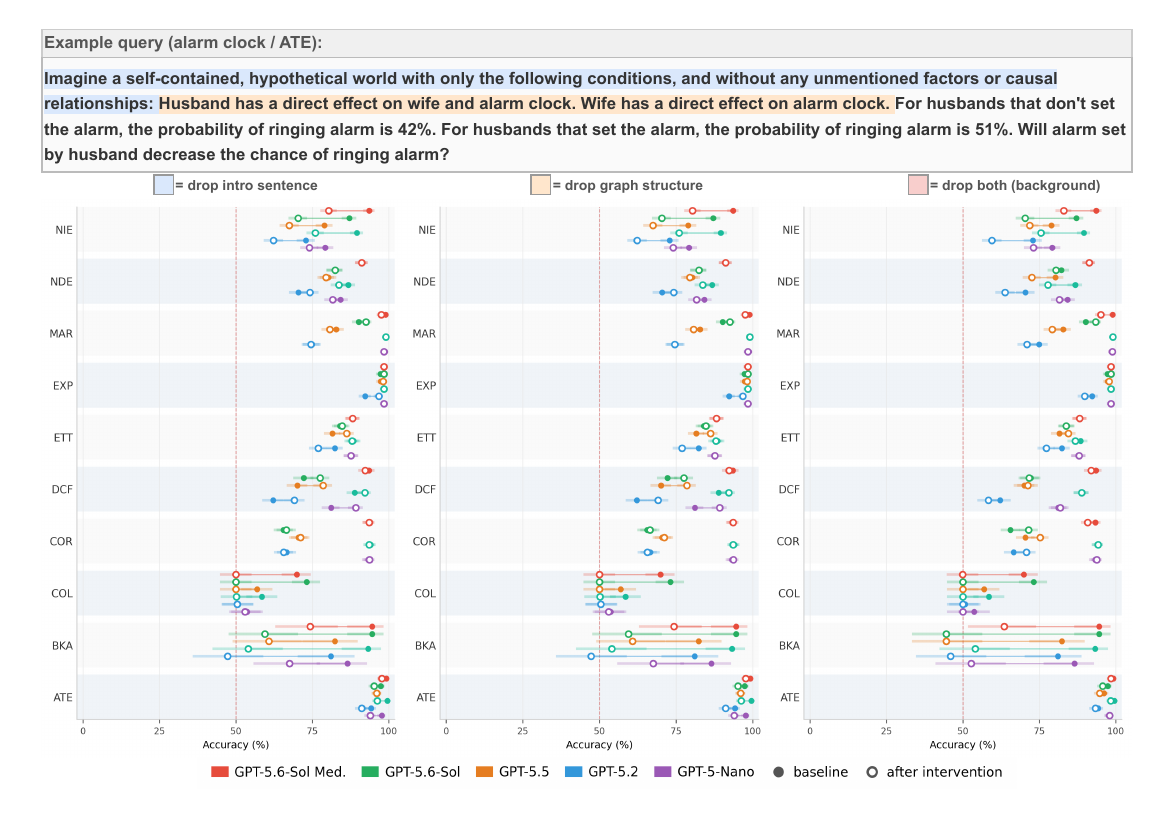}
    \caption{Per-query-type model accuracy under interventions that remove elements of the queries that are critical for formal causal reasoning. We show 95\% Wilson CI.}
    \label{fig:intervention_1A}
\end{figure}

\begin{figure}[!htbp]
    \centering
    \includegraphics[width=1\linewidth]{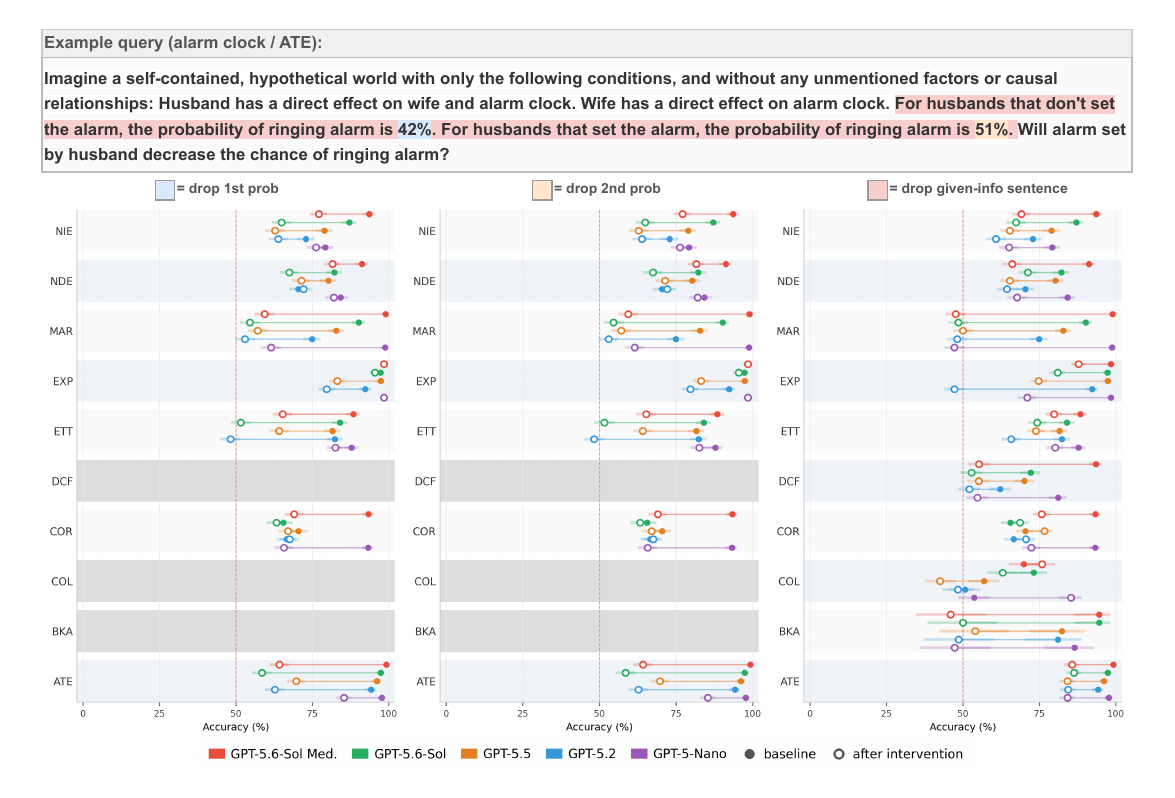}
    \caption{Per-query-type model accuracy under interventions that remove elements of the queries that make it impossible to calculate the answer. We show 95\% Wilson CI.}
    \label{fig:intervention_2A}
\end{figure}

We abbreviate query types present in \CLADDER as follows: Marginal (MAR), Correlation (COR), Back-door Adjustment (BKA), Average Treatment Effect (ATE), Effect of Treatment on the Treated (ETT), Natural Indirect Effect (NIE), Natural Direct Effect (NDE), Deterministic Counterfactual (DCF), Explaining Away (EXP), and Collider Bias (COL). Figures \ref{fig:intervention_1A} and \ref{fig:intervention_2A} show the per-query-type accuracy change for the six interventions that were discussed in Section \ref{sec:model-high-perf-essential}.

\newpage
\subsection{Sources of Determinism in \CLADDER Data}
\label{app:sources-of-determinism}

To understand the mechanisms underlying model performance on \CLADDER, we conducted a systematic analysis of the dataset's generation codebase. This analysis reveals three distinct levels at which answer determinism is introduced into the benchmark, each of which provides a potential shortcut a model could exploit without performing genuine causal reasoning.

\subsubsection{Parametric determinism}
\CLADDER's \texttt{DifficultyBuilder} generates questions by sampling Structural Causal Model parameters from predefined specification lists stored in per-story configuration files. In the easy split, these specifications assign consistent sign directions to causal mechanisms across all 37 non-nonsense stories, implying the direction of the causal effect can be typically determined by reading a single mechanism edge in the prompt, without computing anything. The hard split systematically reverses these signs. In both cases, knowing which story a query comes from is often sufficient to predict the correct answer regardless of the specific probability values provided.

\subsubsection{Structural determinism}
The collision graph topology ($X \rightarrow V \leftarrow Y$), Y has no parents and is therefore statistically independent of X by construction. Any intervention on X cannot affect Y, so the average treatment effect $E[Y|do(X=1)] - E[Y|do(X=0)] = 0$ for every parameter setting. All queries pairing a collision graph with an ATE, ETT or COR query type therefore have a predetermined answer that can be read from the graph structure.

\subsubsection{Formula-level determinism}
The COL query type computes the correct answer as $(-1)^{polarity}$, where polarity is a flag that controls whether the question is phrased as "does X affect Y" (answer is always \textit{no}) or "does X not affect Y" (answer is always \textit{yes}). This is independent of the causal graph, the Structural Causal Model parameters, and the numerical correlation value shown in the prompt. The correlation value is computed and displayed to the model, but is never used to determine the ground truth answer.

\subsection{Random Forest Classifier Features}
\label{app:random_forest_classifier}
When training the random forest classifier we use the following features:

\begin{itemize}

    \item Boolean variable indicating whether the query asks if one quantity is "greater than" or "less than" another quantity.  
    
    \item The number of characters in the query. 
    
    \item The number of probability values that appear in the query. 
    
    \item The first probability that appears in the query. This feature is set to 0 for queries with no probability values.

    \item The second probability that appears in the query. This feature is set to 0 for queries with less than two probability values.

    \item The last probability that appears in the query. This feature is set to 0 for queries with no probability values. Some query types contain 2 or less probability values, so a probability value that appears first or second in the query could also be the last. 
    
    \item Boolean variable indicating whether the first probability value in the query is larger than the second.
    
    \item Boolean variable indicating whether the first probability value in the query is larger than the last.

    \item The product of the first and second probabilities that appear in the query.

    \item The product of the first and last probabilities that appear in the query.
    
    \item The product of the second and last probabilities that appear in the query.

    \item Boolean variables indicating the presence or absence of the following strings: "does it mean," "does not effect," "method,", "case by case," "chance of," "increase," "decrease," "overall," "more likely," "less likely," "when observing," "larger," "smaller," "does the chance of," "does the probability of," "does the likelihood of," "disregard the mediation effect," "positively affect," "negatively affect," "through," "would it be," "instead of," "not," "and," "or," "doesn't," and "correlation."

    \item Boolean variables indicating if the query starts with the words "for" or "would."
    
\end{itemize}

\subsection{Model performance on CounterBench}
\label{app:counterbench}

\begin{table}[!htbp]
\centering
\small
\setlength{\tabcolsep}{6pt}
\begin{tabular}{llc}
\toprule
\textbf{Model} & \textbf{Type} & \textbf{Acc.\ (\%)} \\
\midrule
\multicolumn{3}{l}{\textit{Simple models ($n=1200$)}} \\
\midrule
Random Forest                & Classifier     & $79.1 \pm 2.3$ \\
\midrule
\multicolumn{3}{l}{\textit{Commercial models (one-word prompt, $n=1200$)}} \\
\midrule
GPT-3.5-turbo                & Base           & $48.6 \pm 2.8$ \\
GPT-4.1                      & Base           & $48.8 \pm 2.8$ \\
GPT-4o                       & Base           & $52.2 \pm 2.8$ \\
GPT-5-nano                   & Base           & $51.2 \pm 2.8$ \\
GPT-5.2                      & Base           & $50.8 \pm 2.8$ \\
GPT-5.5                      & Base           & $51.9 \pm 2.8$ \\
GPT-5.5                 & Med.\ reason   & $50.3 \pm 2.8$ \\
GPT-5.6 Sol                  & Base           & $49.3 \pm 2.8$ \\
GPT-5.6 Sol              & Med.\ reason   & $52.1 \pm 2.8$ \\
\midrule
\multicolumn{3}{l}{\textit{Open-weight base models ($n=1200$)}} \\
\midrule
OLMo-3-7B-Instruct           & Base           & $50.4 \pm 2.8$ \\
OLMo-3.1-32B-Instruct        & Base           & $50.4 \pm 2.8$ \\
LLama-3.1-8B-Instruct        & Base           & $50.8 \pm 2.8$ \\
Qwen2.5-3B-Instruct          & Base           & $50.5 \pm 2.8$ \\
\midrule
\multicolumn{3}{l}{\textit{Fine-tuned models, $N=4000$ ($n=1200$)}} \\
\midrule
OLMo-3-7B-Instruct           & Fine-tuned     & $63.0 \pm 2.7$ \\
OLMo-3.1-32B-Instruct        & Fine-tuned     & $71.5 \pm 2.5$ \\
LLama-3.1-8B-Instruct        & Fine-tuned     & $68.9 \pm 2.6$ \\
Qwen2.5-3B-Instruct          & Fine-tuned     & $59.2 \pm 2.8$ \\
\bottomrule
\end{tabular}
\caption{CounterBench Alpha accuracy $\pm$ half-width of Wilson 95\% CI (chance\,=\,50\%). All commercial GPT models perform at or near chance regardless of generation. Fine-tuned open-weight models and the random forest outperform all commercial models.}
\label{tab:counterbench}
\end{table}

\newpage
\subsection{Expanded Token Ablation Experiment}
\label{app:token_ablation}

\begin{figure}[!htbp]
    \centering
    \includegraphics[width=1\linewidth]{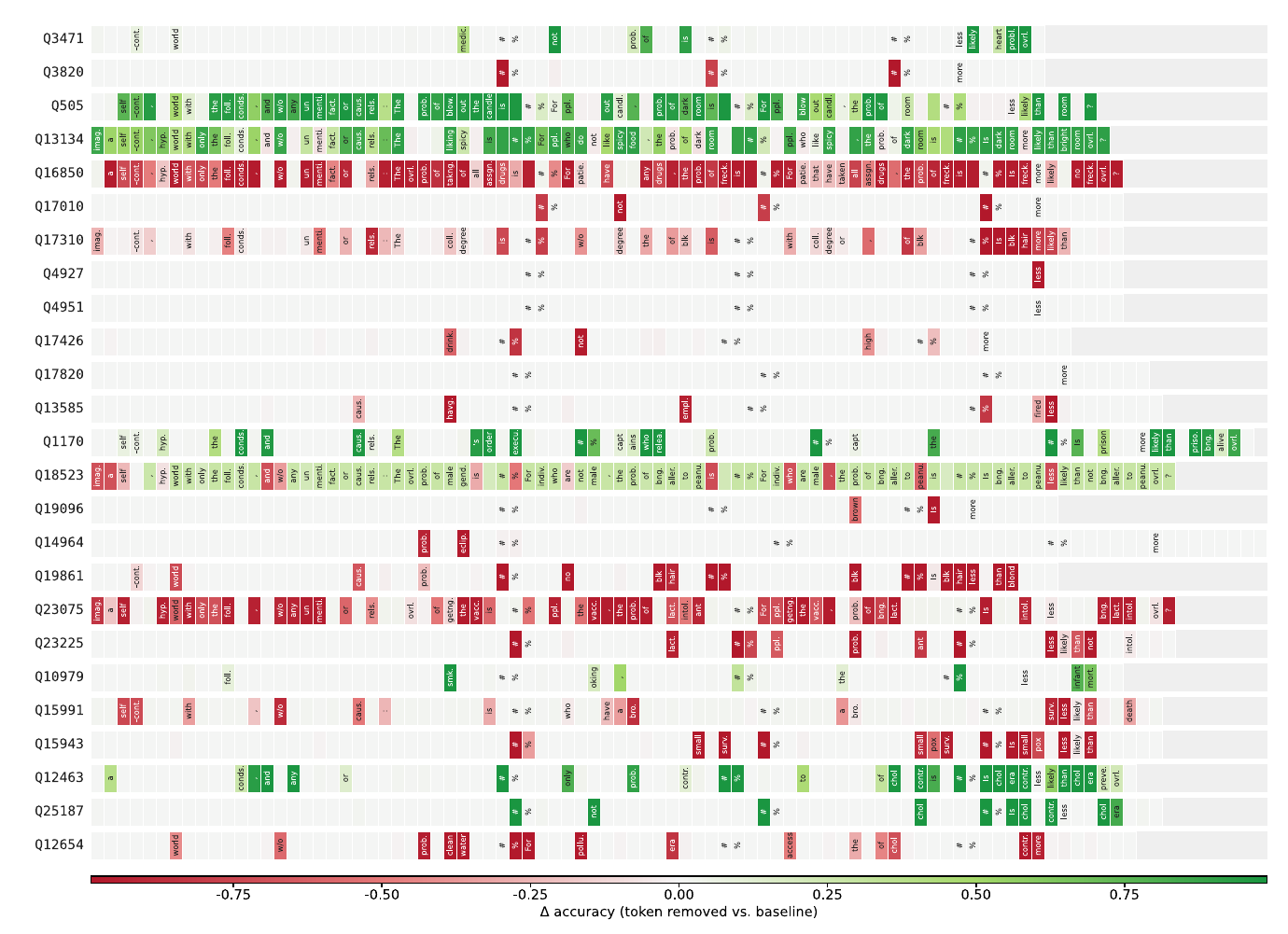}
    \caption{Token ablation analysis on gpt-5.2 for a random sample of 25 evaluated marginal queries. Same visualization as Figure \ref{fig:top-12-ablation}. Color encoding, label conventions, and salience threshold are identical to the main figure.}
    \label{fig:full-token-ablation}
\end{figure}

\end{document}